\documentclass[sigconf]{acmart}
\usepackage{balance}
\AtBeginDocument{%
  \providecommand\BibTeX{{%
    \normalfont B\kern-0.5em{\scshape i\kern-0.25em b}\kern-0.8em\TeX}}}

\copyrightyear{2021}
\acmYear{2021}
\setcopyright{acmcopyright}\acmConference[MM '21]{Proceedings of the 29th ACM International Conference on Multimedia}{October 20--24, 2021}{Virtual Event, China} \acmBooktitle{Proceedings of the 29th ACM International Conference on Multimedia (MM '21), October 20--24, 2021, Virtual Event, China}
\acmPrice{15.00}
\acmDOI{10.1145/3474085.3475269}
\acmISBN{978-1-4503-8651-7/21/10}
\begin{document}

\settopmatter{printacmref=true}
\fancyhead{}
\title{MV-TON: Memory-based Video Virtual Try-on network}

\author{Xiaojing Zhong}
\authornote{indicates equal contribution.}
\affiliation{%
  \institution{School of Software Engineering}
  \institution{South China University of Technology}
  \city{Guangzhou}
  \country{China}}
\email{vzxj@mail2.gdut.edu.cn}

\author{Zhonghua Wu}
\authornotemark[1]
\affiliation{%
\institution{S-lab for Advanced Intelligence}
\institution{School of Computer Science and Engineering}
  \institution{Nanyang Technological University}
  \city{}
  \country{Singapore}
}
\email{zhonghua001@e.ntu.edu.sg}

\author{Taizhe Tan}
\affiliation{%
   \institution{School of Computers}
 \institution{Guangdong University of Technology}
  \city{Guangzhou}
  \country{China}}
  \email{969313709@qq.com}
  
\author{Guosheng Lin}
\authornote{Corresponding authors: G. Lin (e-mail: gslin@ntu.edu.sg) and Q. Wu (e-mail: qyw@scut.edu.cn).}
\affiliation{%
\institution{S-lab for Advanced Intelligence}
\institution{School of Computer Science and Engineering}
  \institution{Nanyang Technological University}
  \city{}
  \country{Singapore}}
\email{gslin@ntu.edu.sg}

\author{Qingyao Wu}
\authornotemark[2]
\affiliation{%
      \institution{School of Software Engineering}
  \institution{South China University of Technology}
  \city{Guangzhou}
  \country{China}}
\email{qyw@scut.edu.cn}


\begin{abstract}
With the development of Generative Adversarial Network, image-based virtual try-on methods have made great progress. However, limited work has explored the task of video-based virtual try-on while it is important in real-world applications. Most existing video-based virtual try-on methods usually require clothing templates and they can only generate blurred and low-resolution results. To address these challenges, we propose a Memory-based Video virtual Try-On Network (MV-TON), which seamlessly transfers desired clothes to a target person without using any clothing templates and generates high-resolution realistic videos. Specifically, MV-TON consists of two modules: 1) a try-on module that transfers the desired clothes from model images to frame images by pose alignment and region-wise replacing of pixels; 2) a memory refinement module that learns to embed the existing generated frames into the latent space as external memory for the following frame generation. Experimental results show the effectiveness of our method in the video virtual try-on task and its superiority over other existing methods.
\end{abstract}





%% These commands are for a PROCEEDINGS abstract or paper.
\maketitle

\begin{figure}[h]
\centering
\setlength{\abovecaptionskip}{0pt}
\includegraphics[width=0.45\textwidth,height=0.22\textheight]{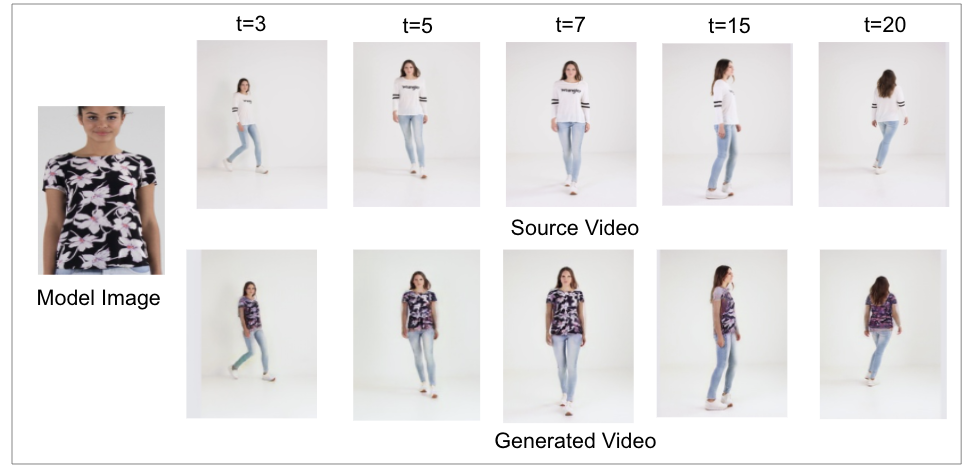}
\caption{An example of virtual try-on with arbitrary videos}
\label{1}
\end{figure}

% \setcopyright{acmcopyright}
% \copyrightyear{2021}
% \acmYear{2021}
% \acmDOI{10.1145/1122445.1122456}
% \acmConference[ACM MM '21]{ACM MM '21: ACM MM '21: ACM Multimedia Conference}{Oct 20--24, 2021}{Chengdu, China}
% \acmBooktitle{Chengdu '21: ACM Multimedia, Oct 20--24, 2021, Chengdu, China}
% \acmPrice{15.00}
% \acmISBN{978-1-4503-XXXX-X/18/06}



\section{Introduction}
Virtual try-on has become an essential task that allows users to try on the desired clothes obtained from social media or shopping websites \cite{wu2020exploring}, which implies such a task can be seen as a type of e-commerce application that offers the convenience of shopping. Currently, with the growth of short video technology, generating high-quality videos of a person wearing desired clothes is of vital importance.
\par 
Previous virtual try-on methods usually focus on image-based operations \cite{han2018viton,wang2018toward,Dong2019TowardsMG}, which are able to try on clothing templates for a target person. Clothing templates are clean product images that do not contain any people, while model images are images of models who wear desired clothes. However, directly applying the image-based methods to video generation may cause the problem of lacking smoothness between the adjacent generated frames. As each frame is regarded merely as an image, the frames of videos cannot mutually facilitate each other. Although \cite{Dong2019FWGANFW} utilizes optical flow to warp the past synthesized frames for following frame generation, it restricts the number of frames to use and overly relies on the results of the preceding generated frames, which may lead to texture misalignment and blurred results. In addition, the requirements for clothing templates in \cite{Dong2019FWGANFW} are typically unavailable and inaccessible. 
\par
In order to address the challenges mentioned above, we propose a two-stage Memory-based Video Virtual Try-On Network (MV-TON). As illustrated in Figure \ref{1}, given a source video and a model image, MV-TON is capable of synthesizing realistic videos that transfer the desired clothes from the model image to the target person in each frame of the video. Specifically, with an intrinsically learned 3d appearance flow, the key idea of the first stage is to employ pose transfer on the model image and further transfer the corresponding regional pixels from the deformed model image to the frame image using a region-wise replacing block in terms of the individual frame.
\par 
However, it is difficult for the first stage to deal with complex scenarios and large deformations due to the non-rigid nature of human bodies. Inspired by the external memory which is normally used in video semantic segmentation \cite{Oh2019VideoOS,Miao2020MemoryAN}, we propose a memory refinement module in which pixel locations are selected from multiple frames to reconstruct space-time information, which can simultaneously retrieve missing pixels and correct wrong pixels. Still, to the best of our knowledge, this work is the first attempt to apply frame aggregation through a memory mechanism to video synthesis. Without additional masks, a dedicated encoder is used to embed the frames generated from the first module into key and value feature maps. To be more specific, key maps address the similarities between the current frame and the past generated frames, as value maps determine the pixels when-and-where to retrieve. Finally, the key and value maps stored in external memory go through a memory read block to reconstruct the intermediate frames. Besides, we employ a flow consistency loss \cite{Lee2019SFNetLO} so that the network can achieve spatio-temporal smoothing during the synthesizing.

\par
To summarize, our MV-TON network comprises two essential modules: 1) a try-on module is responsible for transferring the desired clothes to the target person in each frame of the given video by applying pose alignment and replacing specified pixels, so as to keep prominent appearance details of the person; 2) a memory refinement module aims to combine space-time information of multiple frames with an aggregation mechanism to refine the existing generated frames.
\par 
Due to the insufficiency of genuine paired training data (person videos equipped with the desired clothes), lots of virtual try-on methods \cite{han2018viton, wang2018toward} separate pseudo paired data into various representations to reassemble them, with the input being exactly the same as the output. How to leverage the unpaired data ignored in most approaches is non-trivial. Although the unpaired data has a different appearance with model images, we aim to explicitly control the similarity between their clothes. To this end, we design a novel matching discriminator that enforces the clothes of generated frames to be similar to the clothes of model images.
\par
We note that the evaluation metrics adopted in most virtual try-on tasks do not consider their peculiarities, regarding them as normal image synthesis tasks. Thus we propose a novel evaluation matrix Cycle Transfer Score (CTS). Specifically, obtained an initial output video, we apply the try-on method again to recover clothes from the generated frames to the model images. A robust model should get the low `transfer cost' in the repetitive transformation.

\par
Our contributions can be summarized as below:
\begin{itemize}
    \item {A two-stage framework is carefully designed to overcome the difficulty of dealing with a large variety of poses in video virtual try-on. The proposed memory refinement operation improves the details by automatically correcting what is wrong or missing in the initial results. The optical flow of the generated videos is regularized by a flow consistency loss. }
    
    \item {We propose a matching discriminator that leverages the unpaired data to better train the model with a hybrid training strategy.} 
    
    \item {We propose an evaluation metric Cycle Transfer Score (CTS) specially designed for virtual try-on tasks to evaluate the objective quality of generated videos.}
\end{itemize}

\begin{figure*}
\centering
\setlength{\abovecaptionskip}{0pt}
\includegraphics[width=0.75\textwidth,height=0.35\textheight]{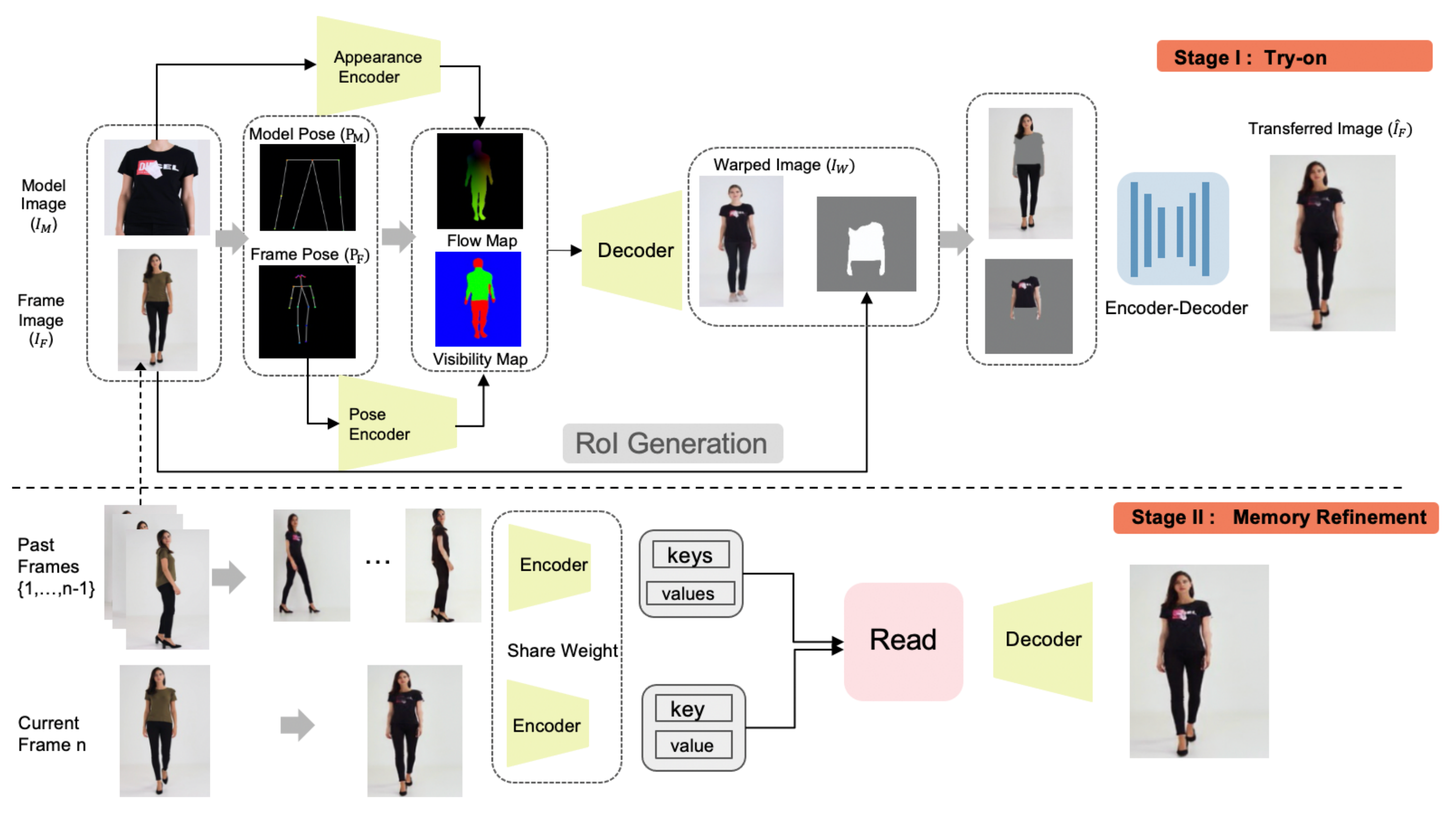}
\caption{Overview of our MV-TON framework. Our proposed MV-TON has two main stages. In the first stage, given a model image $I_M$ and an arbitrary frame image $I_F$, MV-TON can achieve the pose transformation according to their poses extracted by a pose estimator to get the warped image $I_W$. Then, the transferred image $\hat{I}_F$ can reserve the appearance of $I_F$ except for clothing items and arms to fuse with the clothes extracted from $I_W$. The results generated from stage I are further refined by the second stage that consists of a memory refinement module which memorizes the features of past frames to strengthen the current frame.}
\label{2}
\end{figure*}

\section{RELATED WORKS}
{\bf Person Image synthesis.} With the development of Generative Adversarial Networks (GANs) \cite{Goodfellow2014GenerativeAN}, person image synthesis has made significant progress through adversarial learning. \citet{Ma2018DisentangledPI} proposed a disentangled representation to generate person images with arbitrary poses. \cite{Siarohin2018DeformableGF,Albahar2019GuidedIT,Ma2017PoseGP,Isola2017ImagetoImageTW} utilized conditional Generative Adversarial Networks (cGANs) \cite{Mirza2014ConditionalGA} to translate skeleton poses into person images. \cite{Ma2017PoseGP} adopted a variant of the conditional DCGAN \cite{Radford2016UnsupervisedRL} architecture that conditioned both on a reference image and a specified pose. \citet{Li2019DenseIA} proposed a network that learns the mapping from an input skeleton pose to a target so as to guide the feature warping. In contrast to the above methods, this work handles the transformation with an intrinsic 3D appearance flow since the identity mapping is hard to learn directly from images. 
\par
{\bf Virtual Try-on.} 3D try-on methods are mostly based on computer graphics. \citet{Guan2012DRAPEDA} first allowed 3D human bodies in any pose to be dressed in 2D clothing. \citet{Sekine2014VirtualFB} proposed a virtual fitting approach for adjusting 2D clothing images on users by modeling their body shapes from depth images. The principles of \cite{PonsMoll2017ClothCapS4} were used to capture the garment geometry in motion on 3D models. However, to avoid the time-consuming raised by 3d model rendering, \cite{han2018viton,wang2018toward,Hsieh2019FashionOnSI,Yu2019VTNFPAI,Dong2019TowardsMG} directly synthesized a photo-realistic image by transferring the desired clothes onto the person, which is computationally efficient. \citet{han2018viton} used shape context to align features through clothing-agnostic person representations. \citet{wang2018toward} described an improved way of \cite{han2018viton} which learns to estimate the transformation parameters. \citet{Dong2019TowardsMG} proposed a framework that adopts decomposed binary masks to predict human parsing maps \citet{wu2019keypoint} so as to guide the generation of person images. In contrast to the above methods requiring clothing templates, \citet{Wu2019M2ETryON} directly used available model images from which clothing pixels could be sampled and transferred to the person's images. Two branches were applied separately in \cite{Raj2018SwapNetIB} to disentangle the body shape and appearance information, which can be reassembled as desired. There are few explorations in the video-based virtual try-on. \citet{Dong2019FWGANFW} exploited optical flow from pose sequences to warp the preceding frame and used a warping module to warp clothes simultaneously. Nevertheless, \cite{Brox2004HighAO} determines the upper limit of the correctness of the flow used in \cite{Dong2019FWGANFW}, and this method didn't take full advantage of whole frames since it utilized a few past frames to synthesize the future frame.
\par 
{\bf Memory Mechanism. }  Many previous methods for document Q$\&$A\cite{Miller2016KeyValueMN,Kumar2016AskMA} embedded memory information into key and value feature vectors individually. Recently, memory mechanisms have been applied in diverse domains. For instance, \citet{Yang2018LearningDM} proposed a dynamic memory network to retrieve the related template for matching objects in the visual tracking task. \citet{Na2017ARM} first introduced a network using multi-layer convolutions applied to writing and reading in the movie understanding task. \citet{Oh2019VideoOS} utilized a space-time memory mechanism to store informative knowledge.

\section{Method Description}
\subsection{Problem Formulation and Notations}
Formally, given a person video $\textbf{V}=\{I_{F}^{i}\}^{N}_{i=1}$, where N is the number of the frames of the video, and a model image $I_M$, we aim to generate a spatial-temporal coherent realistic video \textbf{O}, the person of which wears the desired clothes the same as $I_{M}$. Let \textbf{P}=$\{P_{F}^{i}\}^{N}_{i=1}$ denote the pose sequence extracted by the Human Pose Estimator \cite{Cao2017RealtimeM2}, which estimates 18 joints of each human body. $P_{M}$ means the pose of $I_{M}$. Our pipeline is a two-stage architecture designed for individual frame try-on and multiple frame aggregations. Our goal is to learn the mapping of $(I_{M},P_{M},I_{F}^{i},P_{F}^{i})\to \hat{I_{F}^{i}}$ with respect to the i-th frame and further $\hat{V} \to \textbf{O} $ where $\hat{V}=\{\hat{I}_{F}^{i}\}^{N}_{i=1}$.
\subsection{Generator}
\subsubsection{Try-on Module}
The try-on module aims to transfer the specified region-wise pixels from model images to target frames, including clothes and arms, which reserves the remaining appearances of the person to the greatest extent possible.
\par
As depicted in the upper of Figure \ref{2}, an appearance encoder and a pose encoder are used to encode $I_{M}$ and $P_{F}^{i}$ in multi-scales for the current processing frame, respectively. Inspired by \cite{Li2019DenseIA}, we estimate the intrinsic 3D appearance flow between ($P_{F}^{i},P_{M}$) by modeling SMPL \cite{Kanazawa2018EndtoEndRO} in the 3D space for human pose transfer. The parts of self-occlusion and out-of-plane are distinguished from the image through learning a visibility map. With a spatial transformer layer, the flow warps the feature maps at different scales obtained above and then divides them into exclusive regions according to the visibility map. The warped feature maps are decoded to construct the image ${I_{W}}$ which has the same pose as $I_{F}^{i}$. We subsequently introduce a region replacing block to transfer the specified region-wise pixels from ${I_{W}}$ to $I_{F}^{i}$. 

\begin{figure*}[t]
\centering
\setlength{\abovecaptionskip}{0pt}
\includegraphics[width=0.7\textwidth,height=0.27\textheight]{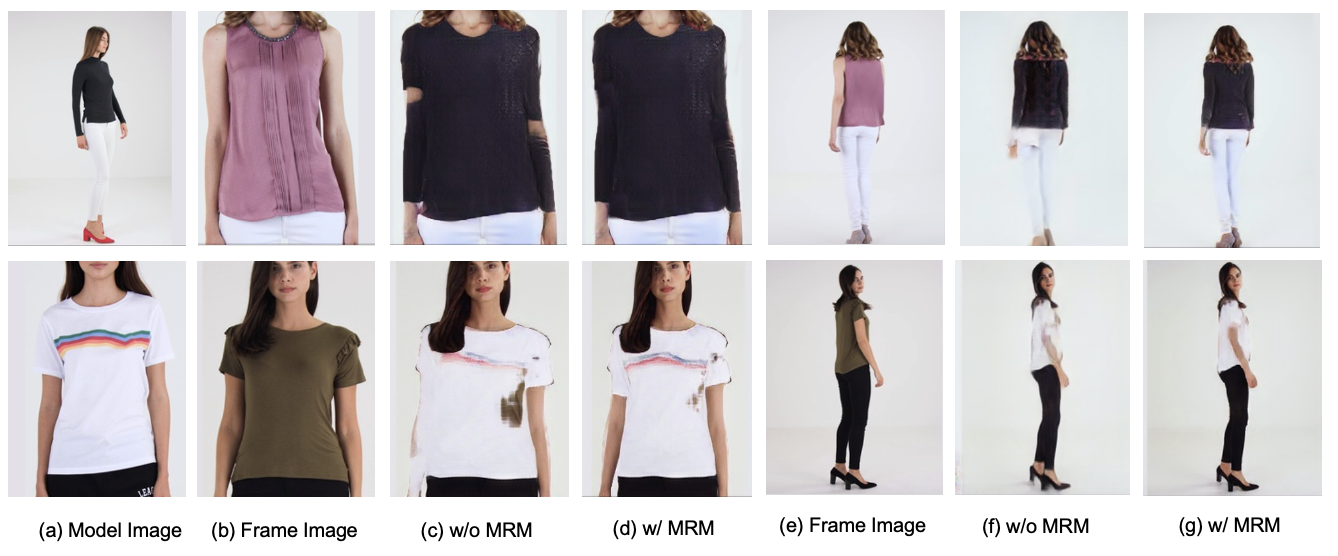}
\caption{\textbf{Effects of the memory refinement module(MRM). (a) A model image.(b)(e) Two different frames randomly selected from a video. (c)(f) The results generated without MRM. (d)(g) The results generated with MRM. }}
\label{3}
\end{figure*}

\par
\rm{\textbf{Region Replacing Block.}}
\rm{With }the warped model image $I_{W}$ in the target pose $P_{F}^{i}$, we use the pre-trained CIHP \cite{Gong2018InstancelevelHP} to generate the human parsing map S and subsequently convert S into 20 channel heat maps $\textit{M} \in R^{20 \times H \times W}$. Each channel is a binary mask and represents one class of person's parts. 
To extract the clothing and arm pixels, we train a Region of Interest (RoI) generation network that considers a full body mask $\textit{M}$ as pseudo ground truths. The decomposed components are computed by multiplying the image with corresponding masks. The region $\textit{M}_{C} \in R^{3 \times H \times W}$contains clothes, right arms, left arms. Our merged component $\hat{M}$ can be expressed as below:
\begin{equation}
\hat{M}={\textit{F} _\bigtriangledown((M_{C} \odot I_{W}) \oplus (I_{F}\odot(1 - M_{C})))},
\end{equation}
where $\odot$ indicates element-wise multiplication for each pixel, and $\oplus$ means channel dimension concatenation. $F_\bigtriangledown(\theta)$ is an encoder-decoder network aiming to fit two individual parts into one.

\par
This module can be used to transfer not only clothes but also other components of dressed people. We observed that faces were prone to losing during the generation \cite{Li2019DenseIA,Dong2019FWGANFW}. It can be solved through replacing the facial region ${\textit{M}_{F} \in R^{3 \times H \times W}}$ from the ultimate synthesized image $\hat{I}$ to the corresponding pixels of $F_{I}$. $M_{F}$ contains faces, necks, and hair. Our merged component $\widetilde{M}$ can be expressed as below:
\begin{equation}
\widetilde{M}={\textit{F}_\bigtriangledown((M_{F} \odot I_{F}) \oplus (\hat{I} \odot (1 - M_{F})))},
\end{equation}
\par
In our experiments, the fitting network $F_\bigtriangledown(\theta)$ can fit two arbitrary regions into a realistic-looking image with only one training.
\subsubsection{Memory Refinement Module}
\label{subsection:3.2.2}
Our task needs to handle a more challenging scene in which the poses of people are varied, while image scenes merely consider a fixed pose. Therefore, we design a Memory Refinement Module (MRM) that alleviates pixel misalignment and loss by selecting space-time pixel locations to reconstruct the current frame.
\par
As shown in the lower of Figure \ref{2}, $\hat{V}=\{\hat{I}_{F}^{i}\}^{N}_{i=1}$ obtained by try-on module are separated into the past frames and the current frame.
Each past frame and the current frame are fed into an encoder network that shares weight to embed the visual appearance into the latent space. The latent representations are subsequently separated into key and value maps respectively for reading and writing via a memory read block. Besides, if there exists more than one past frame, their key and value maps are concatenated along the temporal dimension as external memory. Key maps focus on addressing the correspondence between aggregated past frames and the current frame. The generated pixels which have inaccurate values are always prone to suppression during the synthesizing since they have low similarities with others. As high frequency details of the appearance are sometimes lost, the value maps of the current frame reconstruct the pixels at feature level, indicating that it is necessary to provide the value maps of the past frames to compensate.

\begin{figure}[t]
\centering
\setlength{\abovecaptionskip}{0pt}
\includegraphics[width=0.4\textwidth,height=0.22\textheight]{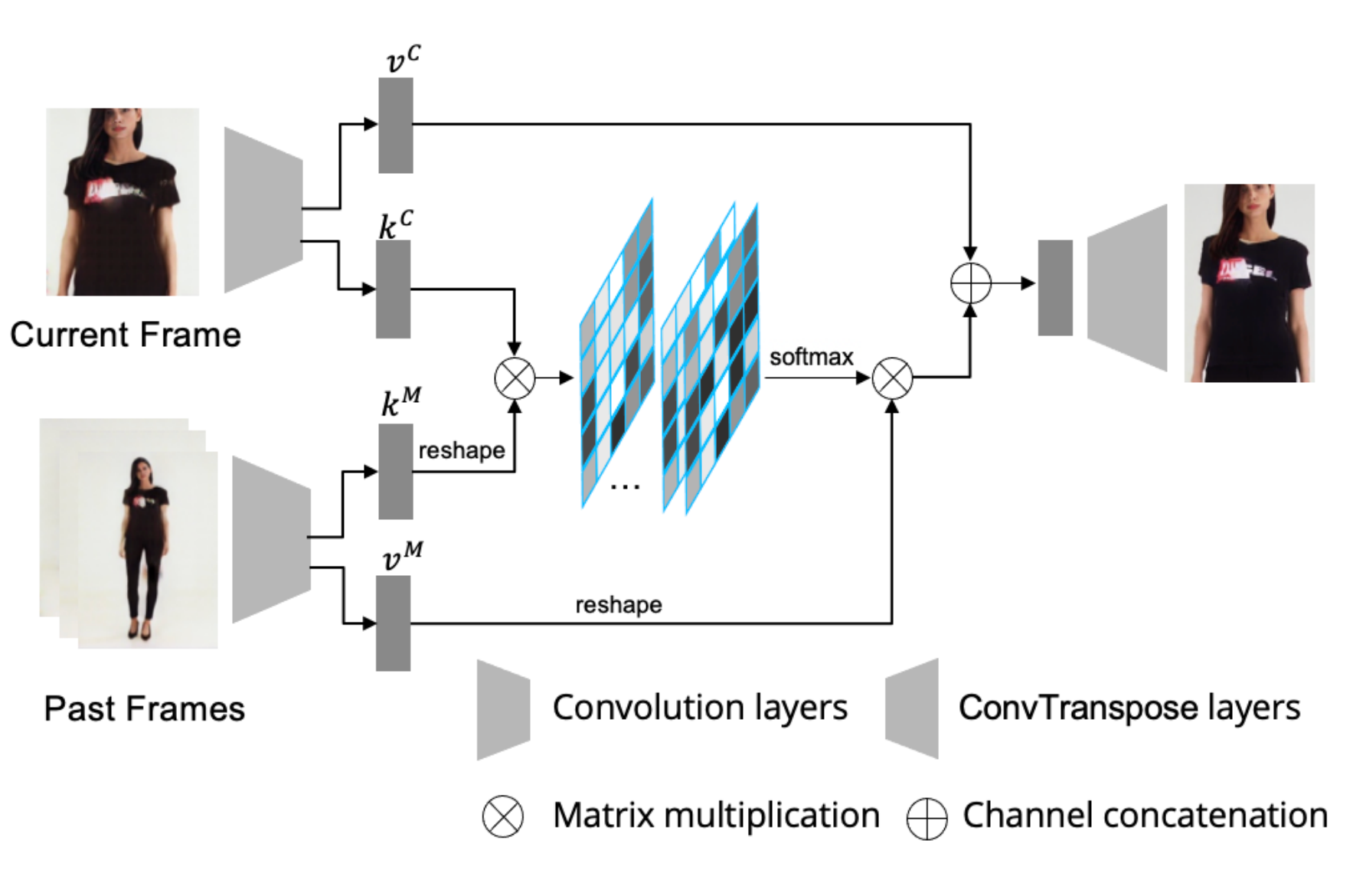}
\caption{\textbf{The architecture of memory read block.} Detailed implementation of the memory read block as are described in Sec.\ref{subsection:3.2.2}.}
\label{4}
\end{figure}

\textbf{Memory Read Block}. As shown in Figure \ref{4}, inspired by \cite{Oh2019VideoOS}, we propose a memory read block. However, instead of using masks and frames together in \cite{Oh2019VideoOS}, we take the video frames as input without additional masks. To be more specific, this block takes $k^{C}$,$v^{C}$,$k^{M}$, $v^{M}$ as input. $k^{M}$ and $v^{M}$ are the aggregations of the key and value maps of the past frames, respectively. $k^{C}$ and $v^{C}$are the encoded key and value maps of the current frame, respectively. The similarity is calculated by comparing every space-time pixel location of the memory key maps with every spatial pixel location of the current key map. The similarity function $\textit{f}$ is defined as follows:
\begin{equation}
f(x_{i},y_{j})= exp(\phi(x_{i}) \cdot \phi(y_{j})),
\end{equation}
where $\textit{i}$ and $\textit{j}$ are the indexes of the spatial pixel locations. $\phi$ denotes the embedding space and $(\cdot)$ denotes dot-product. 
\par
To avoid the variable input size, a normalizing factor  $\frac{1}{Z}$ is necessary, \textit{Z} can be denoted as:
\begin{align}
    \textit{Z}=\sum_{\forall j} f(\textbf{k}_{i}^{C},\textbf{k}_{j}^{M})
\end{align}
\par
Along the dimension $\textit{j}$, it's the same as softmax function by applying $\frac{1}{Z}f(\textbf{k}_{i}^{C},\textbf{k}_{j}^{M})$.
\par
The overall read operation can be summarized as:
\begin{equation}
\textbf{y}_{i}=[\textbf{v}_{i}^{C},\frac{1}{Z}\sum_{\forall j}f(\textbf{k}_{i}^{C},\textbf{k}_{j}^{M})\textbf{v}_{j}^{M}],
\end{equation}
where $[\cdot,\cdot]$ denotes the concatenation.
\par
With the final output of the memory read block, a standard decoder is used to yield the ultimate refined image.
\par
Figure \ref{3} validates the effects of memory refinement module. We can see that this module is indeed able to correct missing and wrong pixels that can not be recovered well by stage I.

\begin{figure}[t]
\centering
\setlength{\abovecaptionskip}{0pt}
\includegraphics[width=0.4\textwidth,height=0.22\textheight]{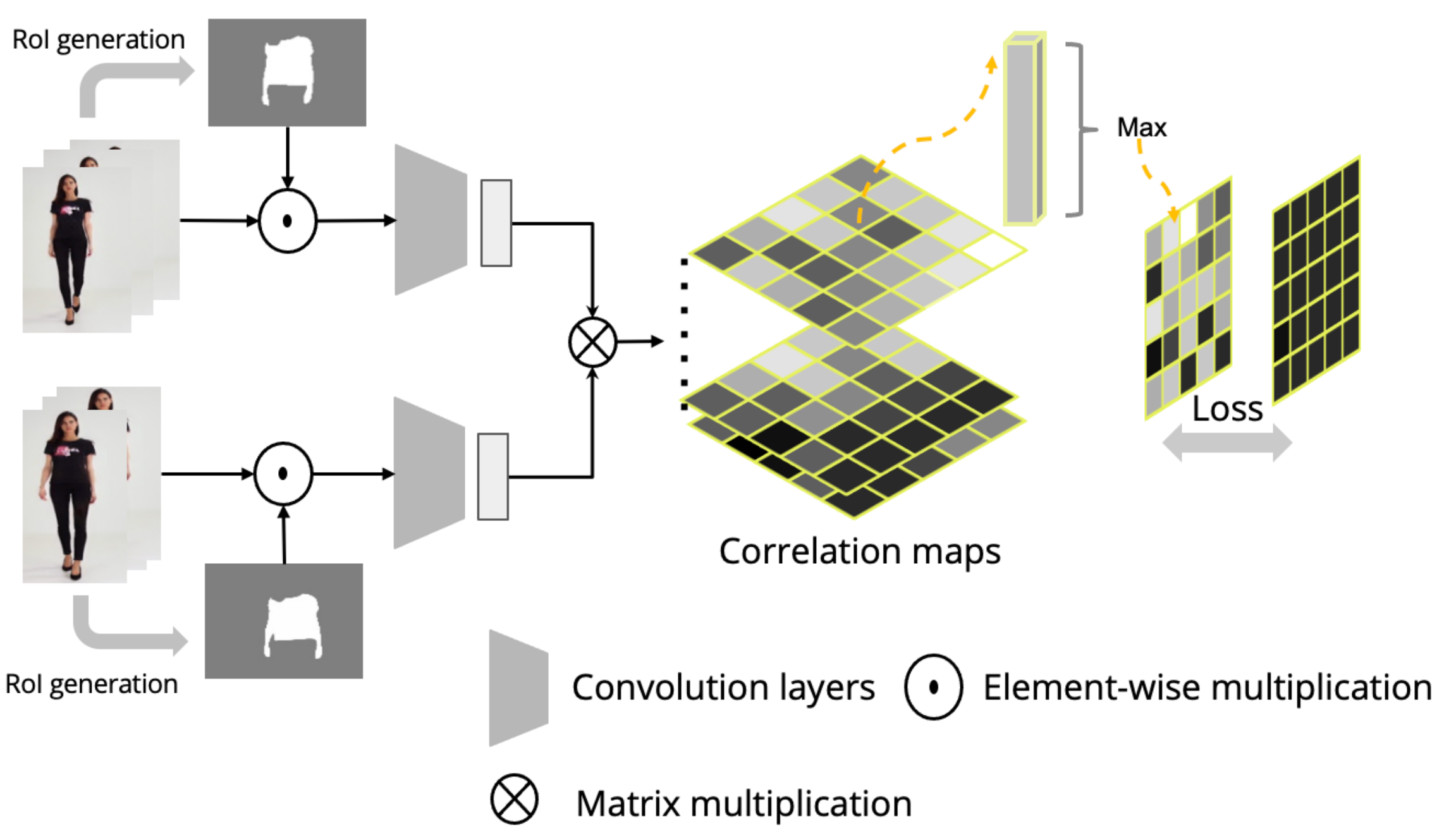}
\caption{\textbf{The process of matching discriminator.} Detailed implementation of the matching discriminator as described in Sec.\ref{section:3.3}.
  }
\label{5}
\end{figure}

\par
\textbf{Hybrid Training.} We cannot acquire paired training data where people wear different clothes but in the exact same pose. Generally, the image itself serves as paired training data ground truth in most image-based virtual try-on methods \cite{han2018viton,wang2018toward,Dong2019TowardsMG}. Training paired and unpaired data simultaneously has proved useful in image synthesis \cite{Tripathy2018LearningIT,Wu2019M2ETryON}. Hence, we take a similar strategy in the memory refinement module.

\subsection{Matching Discriminator}
\label{section:3.3}
Although hybrid training has the potential to contribute towards better results, there are relatively few approaches to supervising unpaired data. Hence, we proposed a novel Matching Discriminator (MD) to force the generation to focus on clothing, which constrains the correlation maps between the specified regions extracted from model images and from transferred images.
\par 
As shown in Figure \ref{5}, the corresponding masks generated by the RoI generation network are sent to extract the clothes to compute correlation maps. Thereafter, for each space-time location, we sample the max value along the dimension. Current discriminators \cite{Wang2018HighResolutionIS,dong2018soft} have been explored on different scales. However, it is hard to incorporate the correspondence between specified regions since the convolutional operation has an effect on whole feature maps. In contrast, we compute an adversarial loss through correlation maps without direct supervision. If the clothes of the synthesized frames are similar to the model images, the result of each patch on the correlation map will be closer to 1, while the opposite will be closer to 0. To make the training more realistic, we randomly sampled two frames of the original video as inputs to calculate their clothing correlation maps.
\par
In more detail, we brief the correlation map as U in the below. Real U is computed as:
\begin{equation}
\textbf{U}_{Real} = \textit{Max}\left[\textit{Conv}(I_{F}^{i}\odot M_{C}) \otimes\textit{Conv}(I_{F}^{j}\odot M_{C}^{'})\right]
\end{equation}
where $\odot$ denotes element-wise multiplication for each pixel and $\otimes$ indicates dot-product. $I_{F}^{i}$ and $I_{F}^{j}$ are the different frames of the original video $\textbf{V}$. $M_{C}$ and $M_{C}^{'}$ are the regional masks, respectively. $\textit{Max}\left[ \cdot\otimes\cdot \right]$ is a function that aims to extract the max value along the specified dimension. \textit{Conv} indicates convolution layers.
Fake U is obtained as:
\begin{equation}
\textbf{U}_{Fake} = \textit{Max}\left[\textit{Conv}({{I}_{F}^{i}}^{'} \odot \hat{M}_{C})\otimes\textit{Conv}(I_{M}\odot \hat{M}_{C}^{'})\right]
\end{equation}
where ${I_{F}^{i}}^{'}$ is the i-th frame of the synthesized videos \textbf{O}. $I_{M}$ is the model image. $\hat{M}_{C}$ and  $\hat{M}_{C}^{'}$ are the regional masks.
\par
\begin{figure}[t]
\centering
\setlength{\abovecaptionskip}{0pt}
\includegraphics[width=0.43\textwidth,height=0.23\textheight]{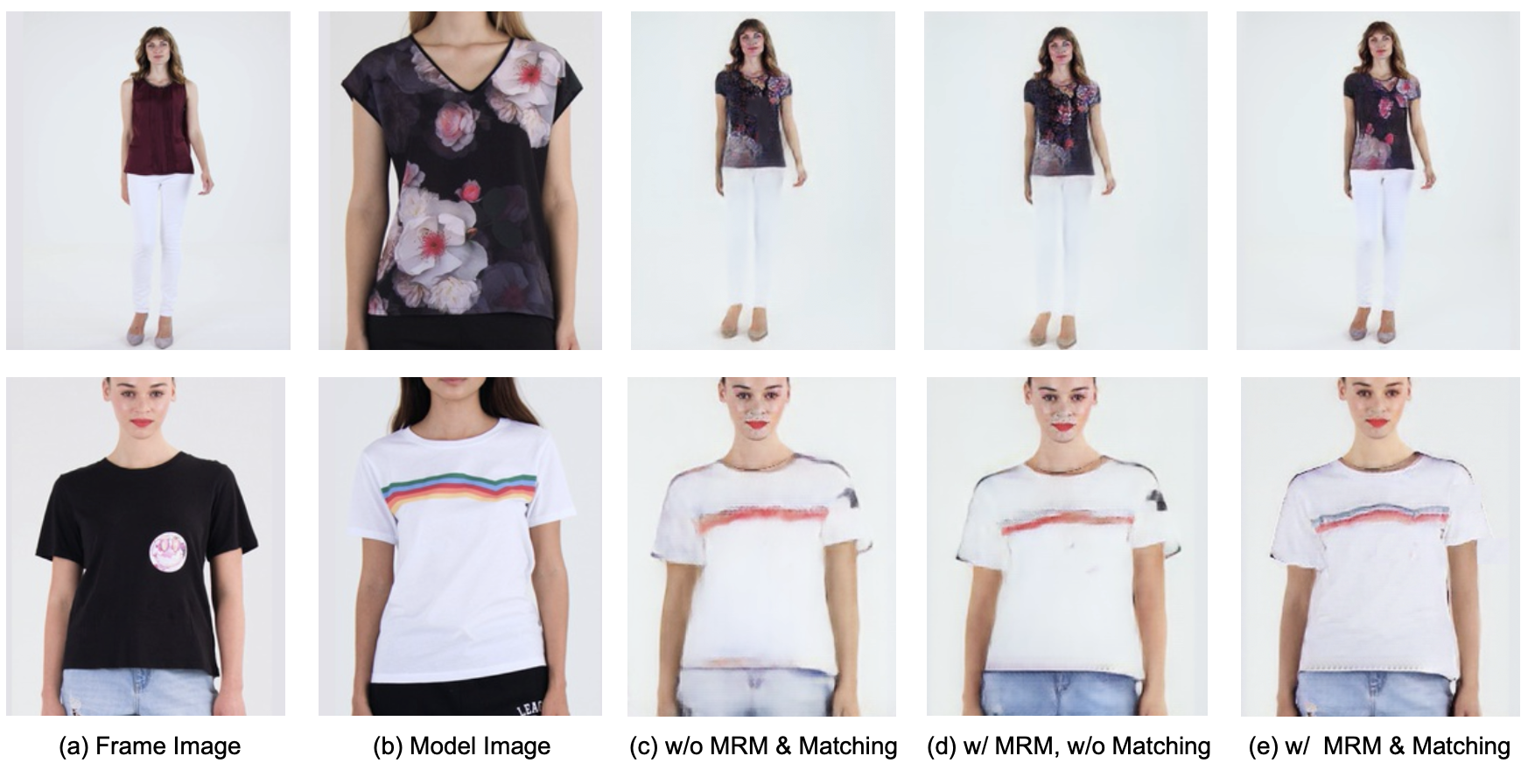}
\caption{\textbf{Effects of the memory refinement module (MRM) and matching discriminator (MD). (a) A frame image randomly selected from videos. (b) A model image. (c) The result generated without either MRM or MD. (d) The result generated without only MD. (e) The result generated with both modules.}}
\label{6}
\end{figure}

The effects of the MD are visualized in Figure \ref{6} (d)(e). When MRM is applied to our model without MD, it may lead to a blurred texture of the results. The matching discriminator enables improved control over the small details. 

\begin{figure}[t]
\centering
\setlength{\abovecaptionskip}{0pt}
\includegraphics[width=0.4\textwidth,height=0.25\textheight]{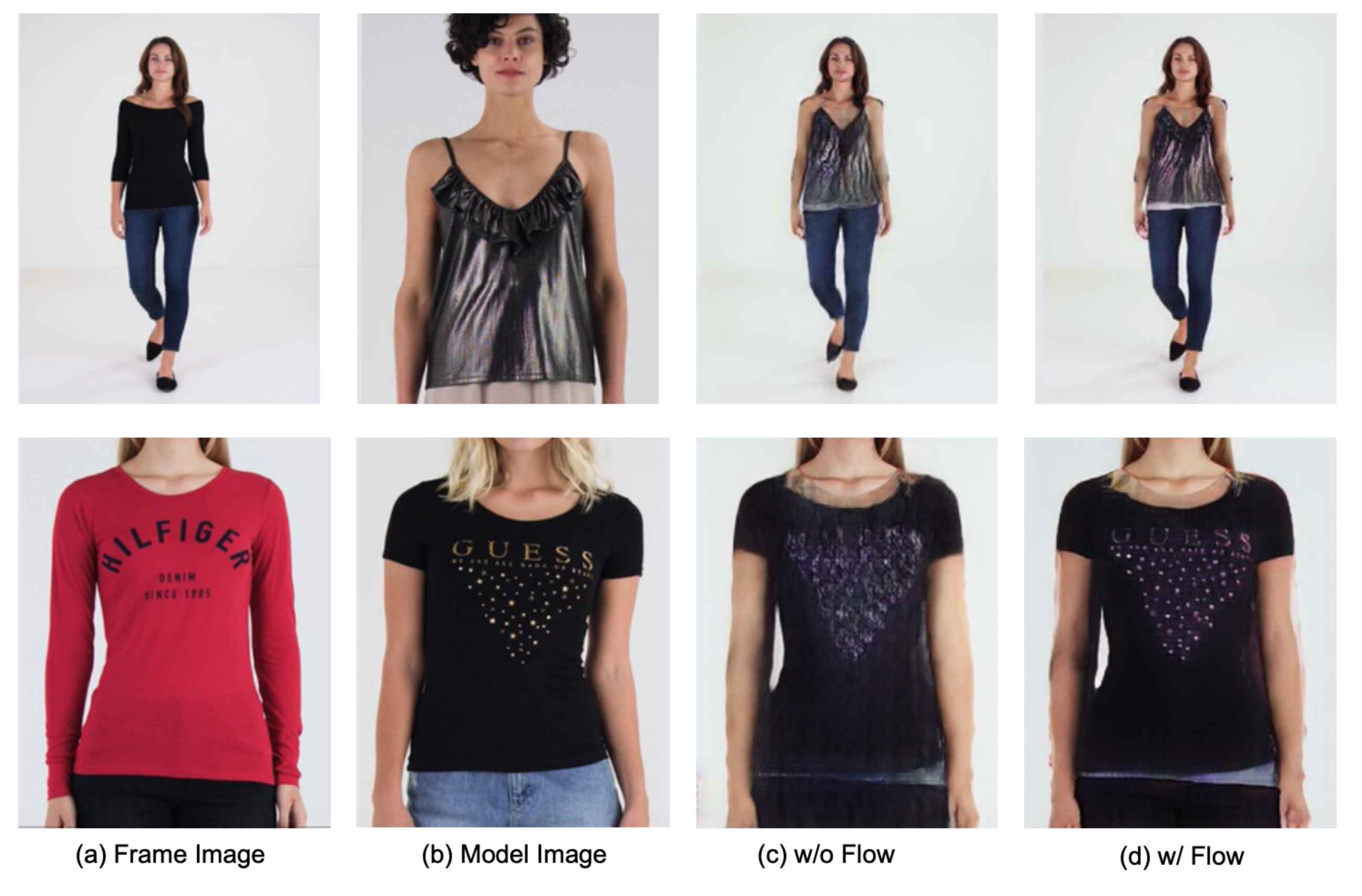}
\caption{\textbf{Effects of the flow consistency loss. (a) A frame image randomly selected from videos. (b) A model image. (c) The result generated without flow. (d) The result generated with flow.}}
\label{7}
\end{figure}

\subsection{Training Losses}
We brief the try-on module as TM and the memory refinement module as MRM. In TM, our goal is to generate photo-realistic images with the desired clothes. Specifically, we apply a vanilla adversarial loss $\mathcal{L}_{adv}$ \cite{Goodfellow2014GenerativeAN}, a L1 reconstruction loss $\mathcal{L}_{L1}$ and a perceptual loss $\mathcal{L}_{perceptual}$ \cite{Johnson2016PerceptualLF} to constrain pose alignment. $\mathcal{L}_{BCE}$ is used to precisely classify the semantic label of each pixel.
The overall objective function $\mathcal{L}_{TM}$ can be computed as:
\begin{equation}
\mathcal{L}_{TM} = \lambda_{1}\mathcal{L}_{adv} + \lambda_{2}\mathcal{L}_{perceptual} + \lambda_{3}\mathcal{L}_{L1} +\lambda_{4}\mathcal{L}_{BCE}
\end{equation}
where hyper-parameters $\lambda_{1},\lambda_{2},\lambda_{3},\lambda_{4}$ are the weights of losses. In our experiments we set $\lambda_{1}$=$\lambda_{4}$=0.01, $\lambda_{2}$=$\lambda_{3}$=1.
\par
In MRM, our goal is to enhance frame consistency and refine the initial results generated from TM. We employ a simplified flow consistency loss \cite{Lee2019SFNetLO} to guide the synthesized videos to keep consistency, which defines a flow field $\mathcal{F}^s$ from the preceding frames to the current frame. $\mathcal{F}^t$ represents the opposite direction.
$\mathcal{L}_{flow}$ can be defined as:
\begin{equation}
    \mathcal{L}_{flow} = \frac{1}{|N|}\sum_{\textit{p}}||(\mathcal{F}^{s}(\textit{p})+\mathcal{F}^{t}(\textit{p}))\odot M(\textit{p})||^{2}_{2}
\end{equation}
where p denotes a pixel location. $\textit{M}$ is the foreground mask of the frame. N is the number of the foreground pixels in $\textit{M}$. We show the effects of flow consistency loss in Figure \ref{7}, which enables the generated videos to better keep consistency.

\par
For paired data, inspired by Pix2PixHD \cite{Wang2018HighResolutionIS}, we adopt a multi-scale GAN to calculate the adversarial loss $\mathcal{L}_{mGan}$. Besides $\mathcal{L}_{flow}$ and $\mathcal{L}_{mGan}$, we also apply a L2 mean squared Loss $\mathcal{L}_{L2}$, $\mathcal{L}_{L1}$ and $\mathcal{L}_{perceptual}$ to tackle the challenge of accurate synthesis.
The overall objective function $\mathcal{L}_{MRM_{pair}}$ can be computed as:
\begin{equation}
\begin{aligned}
    \mathcal{L}_{MRM_{pair}} =  \gamma_{1}\mathcal{L}_{L1}+\gamma_{2}\mathcal{L}_{L2}+\gamma_{3}\mathcal{L}{_{perceptual}}\\
    +\gamma_{4}\mathcal{L}_{flow}+\gamma_{5}\mathcal{L}{_{mGan}}
\end{aligned}
\end{equation}
where hyper-parameter $\gamma_{i}$, i$\in\{1,2,3,4,5\}$ are the weights of losses. In our experiments we set $\gamma_{1}$=$\gamma_{2}$=$\gamma_{5}$=1, $\gamma_{3}$=$\gamma_{4}$=0.1.

With the matching discriminator described in Sec.\ref{section:3.3}, $\mathcal{L}_{match}$ can be formulated as below:
\begin{equation}
\begin{aligned}
\mathop{min}\limits_{G}\mathop{max}\limits_{D}\mathcal{L}_{match}(G,D)=\textit{E}\left[logD(V)\right]\\
+\textit{E}\left[log(1-D(G(\hat{V}),M_{I}))\right]
\end{aligned}
\end{equation}
where V denotes the original videos and $\hat{V}$ indicates the videos generated by TM. $I_{M}$ means the model images. 

The overall objective function $\mathcal{L}_{MRM_{unpair}}$can be calculated as:
\begin{equation}
\mathcal{L}_{MRM_{unpair}} = \beta_{1}\mathcal{L}_{flow} + \beta_{2}\mathcal{L}_{match}
\end{equation}
where hyper-parameters $\beta_{1},\beta_{2}$ are the weights of losses. In our experiments we set $\beta_{1}$=0.1, $\beta_{2}$=1.

\section{Cycle Transfer Score}
\label{section:4}
\begin{figure}[t]
\centering
\setlength{\abovecaptionskip}{0pt}
\includegraphics[width=0.35\textwidth,height=0.25\textheight]{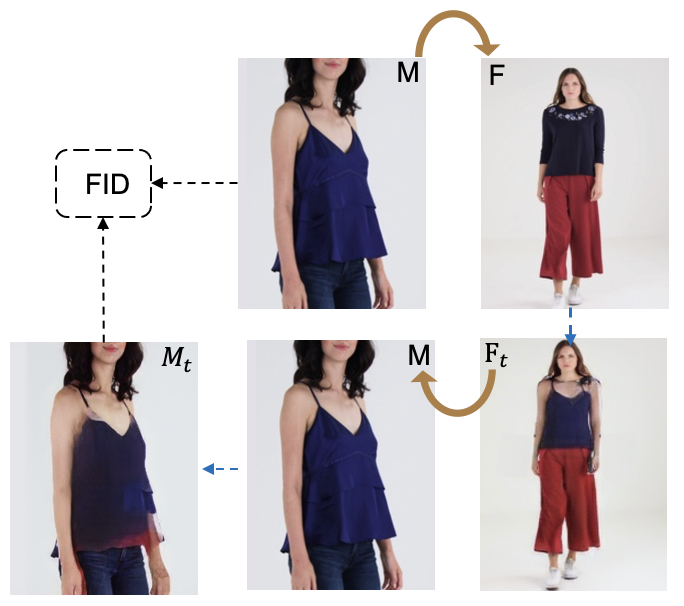}
\caption{\textbf{The process of calculating cycle transfer score. It is composed of two steps as one cycle, and uses FID to evaluate the quality of a cycle.}}
\label{8}
\end{figure}

Among the existing methods, the most widely used evaluation metrics are designed for ordinary synthesis tasks, which ignore the characteristics of virtual try-on. Performing transfer in a cyclic manner, the clothes should keep most of the details so that the model can prove robust in the challenge of pose transformation. In light of this, we propose a novel evaluation metric \textbf{Cycle Transfer Score (CTS)} specially designed for virtual try-on tasks.
Figure \ref{8} demonstrates a cyclic transfer process. 
First, we apply the target virtual try-on method to transfer an arbitrary component from the model image $\textit{M}$ to the original image $\textit{F}$. Then we apply the same method to perform the transfer in an opposite direction -- we transfer from image $\textit{F}_{t}$ to image $\textit{M}$ to generate $\textit{M}_{t}$. The result $\textit{M}_{t}$ is expected to have a similar appearance as $\textit{M}$. 
We measure the difference between $\textit{M}$ and $\textit{M}_{t}$ as the cycle transfer score to reflect the level of cyclic consistency for a given virtual try-on method.
In particular, any kind of pixel level or feature level loss function can be used to measure such a difference, such as mean squared loss and perceptual loss. In our experiment, we compute CTS through Fr\'{e}chet Inception Distance (FID) \cite{Heusel2017GANsTB} which calculates the mean and covariance matrix of features to evaluate image quality.
\par
Notably, besides virtual try-on, CTS can be widely applied to tasks which are desired to transfer the components of images, in order to evaluate the effectiveness of relevant methods.

\begin{figure*}[t]
\centering
\setlength{\abovecaptionskip}{0pt}
\includegraphics[width=0.75\textwidth,height=0.23\textheight]{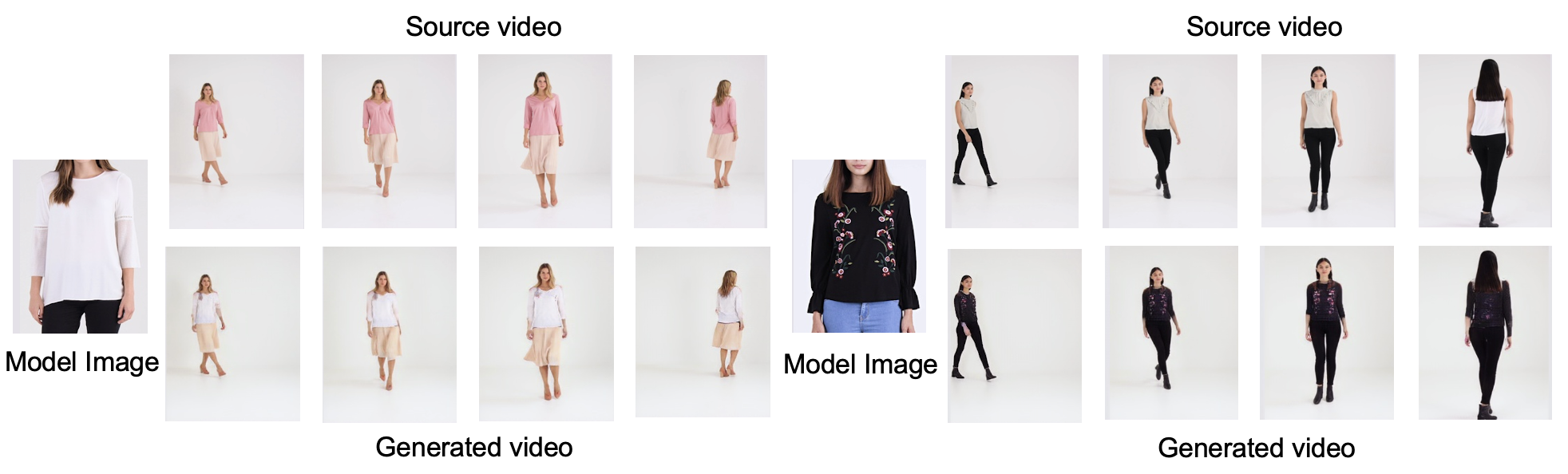}
\caption{\textbf{Results of synthesizing person videos in arbitrary frames.}}
\label{9}
\end{figure*}

\begin{figure*}[t]
\centering
\setlength{\abovecaptionskip}{0pt}
\includegraphics[width=0.6\textwidth,height=0.35\textheight]{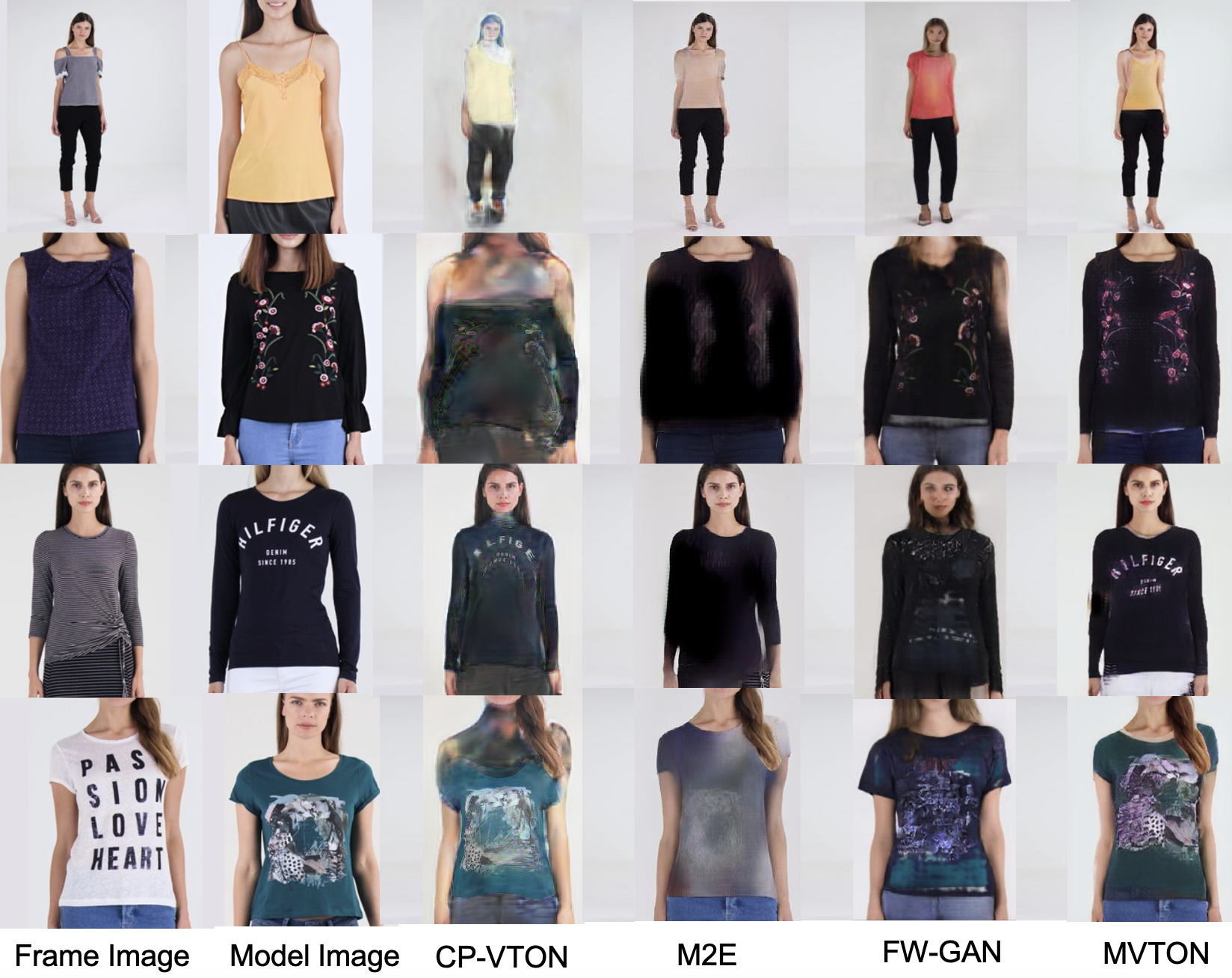}
\caption{\textbf{Qualitative comparisons on the VVT dataset.} As the frame images and model images are inputs, the remaining four columns are the selected synthesized frames generated by different methods. As shown in the picture, MV-TON can better preserve both clothes and appearance.}
\label{10}
\end{figure*}

\section{Experimental Results}
\textbf{Implementation Details}
In stage II, we sample five frames to train due to the limited memory and increase the maximum skipping by five every 20 epochs. We utilize a stochastic gradient solver in two stages with a fixed learning rate of 0.0002 and an adam optimizer ( $\beta_{1}$=0.5 , $\beta_{2}$=0.999). The batch size of stage I and stage II are 8 and 2, respectively. All the experiments were conducted on 2 Nvidia TITAN Xp GPUs.

\par
\textbf{Dataset} We trained and evaluated our MV-TON on the VVT dataset proposed by \cite{Dong2019FWGANFW}, which contains 791 videos of fashion models on the catwalk. We also selected 661 videos for training and 130 videos for testing. Specifically, for stage I, we randomly picked 30 person images of each video in different poses and further combined two of them as paired data, which adds up to 19830 pairs of 661 training videos. In addition, we also picked unpaired images of different videos that have the same number as paired images. For stage II, we crawled 661 person images of whole training sets acted as model images, of which each model image corresponds to one video.

\subsection{Baseline Methods}
We conduct a comparison with the following baseline methods to prove the effectiveness of our proposed MV-TON network. The Characteristic-Preserving Virtual Try-On Network \cite{wang2018toward} (CP-VTON) fits the desired clothes onto the target person. Since the clothing template is not provided, we use an RoI (Region of Interest) generation to extract the clothes from model images. Model to Everyone (M2E-TON) \cite{Wu2019M2ETryON} is similar to our approach, while the former is designed for a fixed pose. We adapt CP-VTON and M2E to our task by assuming each frame to be their user image. Flow-navigated Warping GAN (FW-GAN) \cite{Dong2019FWGANFW} warps the past frames and clothes separately to synthesize the video where the person wears the desired clothes. Similarly, we adopt the RoI generation to extract clothes from model images as the clothing template. Finally, we retrain these approaches on the VVT dataset for a fair comparison.

\begin{figure*}[t]
\centering
\setlength{\abovecaptionskip}{0pt}
\includegraphics[width=0.6\textwidth,height=0.25\textheight]{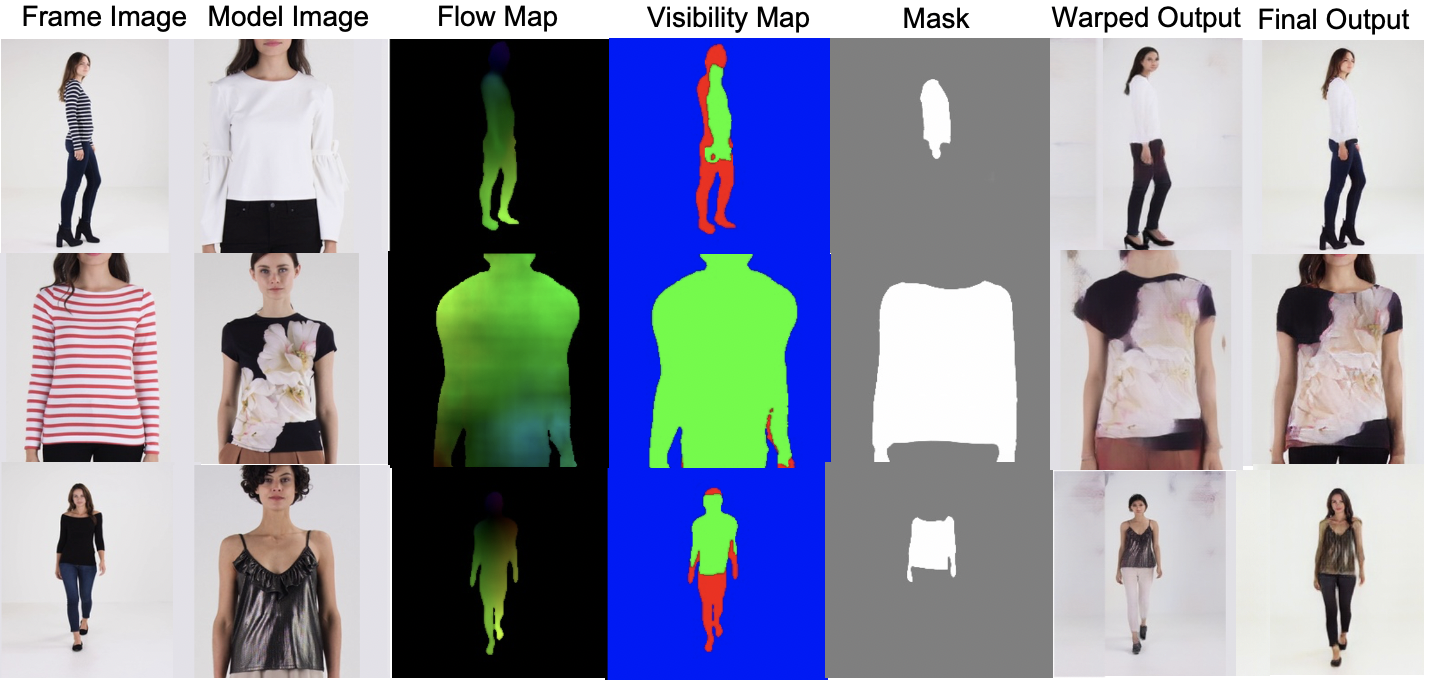}
\caption{\textbf{Intermediate results of MV-TON on the VVT dataset.}}
\label{11}
\end{figure*}

\begin{figure}[t]
\centering
\setlength{\abovecaptionskip}{0pt}
\includegraphics[width=0.3\textwidth,height=0.2\textheight]{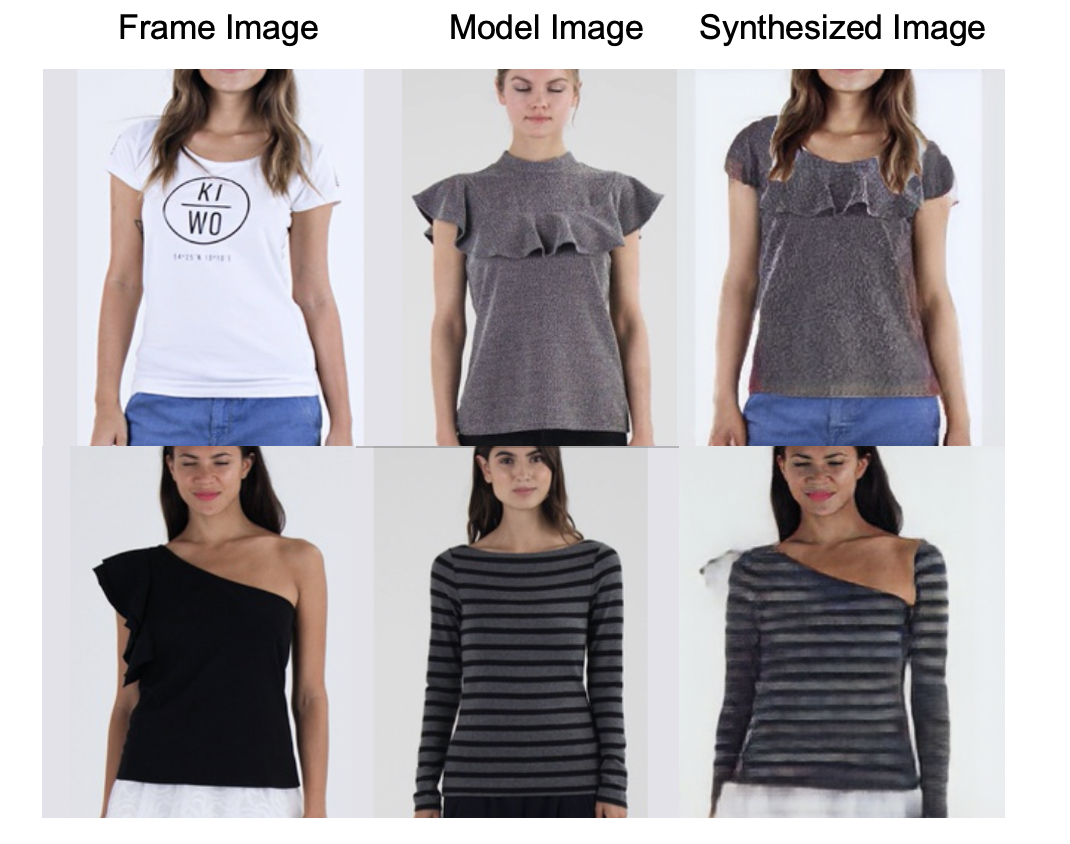}
\caption{Some failure cases of our results which caused by uncommon clothes.}
\label{12}
\end{figure}

\subsection{Qualitative Results}
We show some results of our method in Figure \ref{9} and more video results are available in Supp. Figure \ref{10} presents a visual comparison with the mentioned baseline methods. Compared with other baseline methods, our approach can better preserve both clothes and human appearance. Figure \ref{11} shows the generated intermediate results from different stages of the MV-TON network. Figure \ref{12} shows some failure results of our method caused by the irregular shape of clothes.

\begin{table}
    \centering
    \begin{tabular}{|c|c|c|}
        \hline
         Methods & Cycle Transfer Score\\
        \hline
        \hline
        CP-VTON \cite{wang2018toward}& 270.608 \\
        \hline
        M2E-TON \cite{Wu2019M2ETryON}&135.843 \\
        \hline
        FW-GAN \cite{Dong2019FWGANFW}&241.417 \\
        \hline
        Ours &101.501 \\
        \hline
 
    \end{tabular}
    \caption{Comparison with baseline methods on the VVT dataset.}
    \label{tab:1}
\end{table}
\subsection{Quantitative Results}
\textbf{CTS.} Following \cite{Dong2019FWGANFW}, we choose FID as a measured function, which uses the pre-trained inception model \cite{szegedy2016rethinking} as a feature extractor to exploit desired vectors of the output after a final average pooling layer. Specifically, it computes the mean $\mu$ and covariance matrix $\Sigma$ for the feature vectors from the original model images. It also calculates the same statistics $\Tilde{\mu}$ and $\Tilde{\Sigma}$ for the feature vectors from the synthesized model images. Then the FID can be computed as $||\mu - \Tilde{\mu}||^{2}+ Tr(\Sigma + \widetilde{\Sigma}-2\sqrt{\Sigma\widetilde{\Sigma}})$.

\par
The lower number indicates better performance in Table \ref{tab:1}, demonstrating that our method significantly outperforms the other baseline methods. It also indicates that our method is able to preserve clothing textures in repeatedly synthesizing them.
\par
\textbf{User Study.} We conduct a user study with 30 users to evaluate the quality of the synthesized videos generated by our method and other methods, which use the same reference model image and video sequences. The user is asked to pick one video with better quality of each video group. Meanwhile, the reference model images are shown to make users aware of clothing ground-truth. All samples in the user study are randomly selected from VVT test sets and shown with full resolution. The comparison results of different methods are shown in Table \ref{tab:2}. We can see that the proposed method achieves the best results.

\begin{table}
    \centering
    \begin{tabular}{|c|c|}
        \hline
        Model & User Study$\%$\\
        \hline
        \hline
        CP-VTON \cite{wang2018toward}  & 4.27\\
        \hline
        M2E-TON \cite{Wu2019M2ETryON} & 32.56\\
        \hline
        FW-GAN \cite{Dong2019FWGANFW} &21.02\\
        \hline
        Ours  &42.15\\
        \hline
 
    \end{tabular}
    \caption{User study($\%$)on VVT dataset. The second column represents the wining percentage in the comparison test.}
    \label{tab:2}
\end{table}
\section{Conclusion}
In conclusion, we have presented a network that consists of a try-on module and a memory aggregation module. Our approach transfers the clothes from model images to the target person in videos without using any clothing templates. Extensive evaluations demonstrate our approach performs the best among the existing video-based methods in both qualitative and quantitative. Furthermore, we believe that the proposed approach can be potentially adapted to other video synthesis tasks.
\begin{acks}
This work was supported by National Natural Science Foundation of China (NSFC) 61876208, Key-Area Research and Development Program of Guangdong Province 2018B010108002, and Central Universities of China under Grant D2192860.
This study is supported under the RIE2020 Industry Alignment Fund – Industry Collaboration Projects (IAF-ICP) Funding Initiative, as well as cash and in-kind contribution from the industry partner(s).
This research is partly supported by MOE Tier-1 research grants: RG28/18 (S), RG22/19 (S) and RG95/20.
\end{acks}

\bibliographystyle{ACM-Reference-Format}
\balance
\bibliography{sample-base}

%\appendix

\end{document}